%% file: main.tex
\documentclass[final]{cvpr}

\usepackage{times}
\usepackage{epsfig}
\usepackage{graphicx}
\usepackage{amsmath}
\usepackage{amssymb}
\usepackage{caption}
\usepackage{xcolor}
\usepackage{float}

\DeclareMathOperator*{\argmin}{arg\,min}
\newcommand{\Lagr}{\mathcal{L}}
\newcommand{\loss}{\mathcal{L}}
\newcommand{\pose}{\rho}
\newcommand{\cam}{\mathbf{e}}
\newcommand{\x}{\mathbf{x}}
\newcommand{\z}{\mathbf{z}}
\newcommand{\Iren}{I_{\text{render}}}
\newcommand{\joints}{J}
\newcommand{\joint}{\mathbf{j}}
\newcommand{\jangles}{\Omega}
\newcommand{\jangle}{\boldsymbol\omega}
\newcommand{\weightnorm}{\hat{w}}
\newcommand\Wtilde{\stackrel{\sim}{\smash{W}\rule{0pt}{1.1ex}}}

\usepackage[pagebackref=true,breaklinks=true,colorlinks,bookmarks=false]{hyperref}

\begin{document}

%%%%%%%%% TITLE
\title{Vid2Actor: Free-viewpoint Animatable Person Synthesis from Video in the Wild}
\author{Chung-Yi Weng \quad
Brian Curless \quad
Ira Kemelmacher-Shlizerman \\
University of Washington \\
\vspace{-2ex}
\\
{\small \url{https://grail.cs.washington.edu/projects/vid2actor/}}
}
\twocolumn[{%
\renewcommand\twocolumn[1][]{#1}%
\maketitle
    \begin{center}
    \vspace*{-6ex}
    \includegraphics[width=\textwidth]{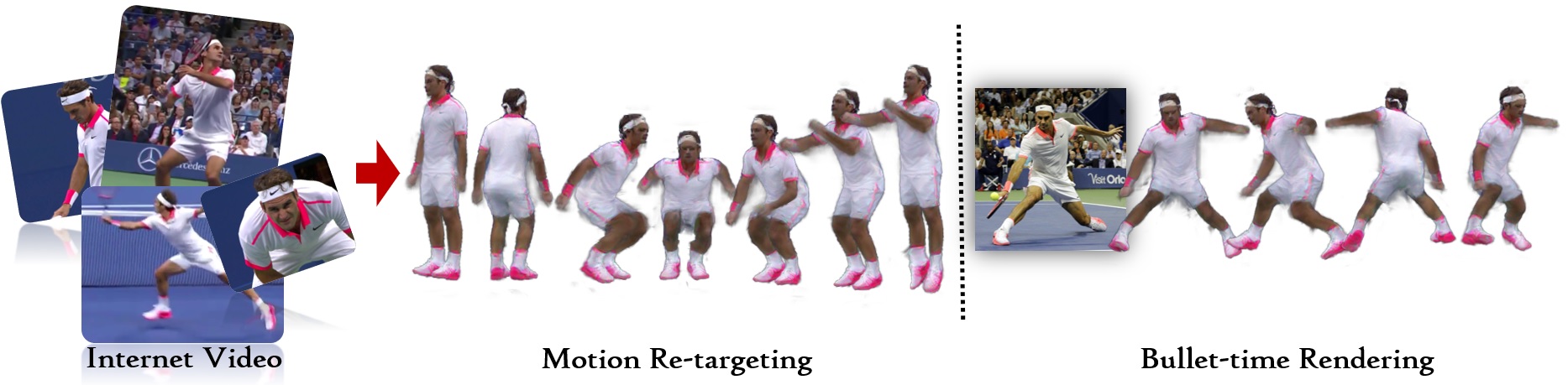}
    \end{center}    
    \vspace{-2ex}
    \captionof{figure}{Given an ``in-the-wild" video, we train a deep network with the video frames to produce an animatable human representation that can be rendered from any camera view in any body pose, enabling applications such as motion re-targeting and bullet-time rendering without the need for rigged 3D meshes. \scalebox{.8}{(AP Photo/Bill Kostroun)} \textbf{Please see all results in the supplementary video:} {\small \url{https://youtu.be/Zec8Us0v23o}.}\\}
    \label{fig:teaser}
}]

%%%%%%%%% ABSTRACT
\begin{abstract}

Given an ``in-the-wild" video of a person, we reconstruct an animatable model of the person in the video. The output model can be rendered in any body pose to any camera view, via the learned controls, without  explicit 3D mesh reconstruction.  At the core of our method is a volumetric 3D human representation reconstructed with a deep network trained on input video, enabling novel pose/view synthesis. Our method is an advance over GAN-based image-to-image translation since it allows image synthesis for any pose and camera via the internal 3D representation, while at the same time it does not require a pre-rigged model or ground truth meshes for training, as in mesh-based learning.  Experiments validate the design choices and yield results on synthetic data and on real videos of diverse people performing unconstrained activities (e.g. dancing or playing tennis). Finally, we demonstrate motion re-targeting and bullet-time rendering with the learned models.

\end{abstract}

%%%%%%%%% BODY TEXT

\input{introduction}
\input{related-work}
\input{representation}

\input{method}

\input{experiment}
\input{conclusion}

{\small
\bibliographystyle{ieee_fullname}
\bibliography{egbib}
}

\end{document}

%% file: introduction.tex
\section{Introduction}

Modeling and rendering humans is a core technology needed to enable applications in sports visualization, telepresence, shopping, and many others.   While a lot of research has been done for calibrated data, e.g.,  multiview laboratory setups \cite{shysheya2019textured, liu2019neural, liu2020neural}, synthesizing any person just from data ``in the wild" is still a challenge. 3D reconstruction approaches require training on synthetic, ground truth meshes, or 3D scans  \cite{saito2019pifu, saito2020pifuhd, alldieck2019tex2shape, huang2020arch, lazova2019360, zhu2020reconstructing}, and image-to-image translation methods focus only on 2D synthesis \cite{chan2019everybody, wang2018video}. 

Our method approaches both challenges by reconstructing a volumetric representation of a person that can be animated and rendered in any pose for any viewpoint.  We take as input a video of the person performing unconstrained activities, e.g., playing tennis or dancing.  The key observation is that image frames from a reasonably long video (minutes to hours) of a person provides a wealth of information about what they look like in different poses and viewpoints, even if each individual pose may only occur once, viewed from one direction.  An important contribution of our paper is the design of a representation and a corresponding deep network that enable generalization to arbitrary poses and views given these images.

Our contributions include a representation of a person as an $RGB\alpha$ volume which can be reposed through a set of volumetric motion weights (adapting the work of~\cite{lombardi2019neural}), the design of a deep network that, after optimizing over the video data, produces the person-specific volumes, a set of losses designed to match the output to both the pixel values and the qualitative appearance of the input frames, and finally, a system that puts these components together to enable fitting to unconstrained videos to produce reposable, free-viewpoint renderings of the person.  We show results on diverse video inputs, outperforming related methods and enabling applications such as motion re-targeting and bullet-time rendering. 

Our approach is not without limitations.  We assume the appearance of the subject is not view-dependent (no specular reflection) and, more significantly, that the appearance does not vary with pose.  The latter assumption means that the results are most faithful to the original if the lighting is fairly diffuse and the shape, e.g., of clothing, does not vary greatly when moving.  In addition, the resolution of results is limited by the voxel grid and by the inherent challenges of registering to, and recovering accurate body poses from, images in the wild.  Despite these limitations, we are able to show convincing results on a variety of sequences which, even if not physically exact, are plausible.

%% file: related-work.tex
\section{Related Work}

\textbf{Image-to-image translation} Recent advances in image-to-image translation \cite{isola2017image} have shown convincing rendering performance on motion retargeting in 2D \cite{chan2019everybody, wang2018high, wang2018video, ma2017pose, esser2018variational, balakrishnan2018synthesizing}. The idea is directly learning a mapping function from 2D skeleton images to rendering output. Those works are usually not robust to significant view changes due to the lack of 3D reasoning. The follow-up works, \cite{shysheya2019textured} and \cite{liu2020neural, liu2019neural} try to overcome this by introducing 3D representations, such as texture maps or pre-built and rigged character models, but both of them need to work on multiple-views laboratory setup. There exists another interesting direction \cite{zhang2020vid2player} that enables retargeting via video sprite rearrangement but still lacks explicit view controls.

\textbf{Novel view synthesis} Novel view synthesis (NVS) is an active area and has been achieved exceptional rendering quality \cite{thies2019deferred, mildenhall2019local, mildenhall2020nerf, srinivasan2019pushing, flynn2019deepview, lombardi2019neural, sitzmann2019deepvoxels, meshry2019neural, li2020crowdsampling, thies2019deferred} recently. In particular, our work is highly related to the methods that uses volumes as intermediate representations \cite{mildenhall2020nerf, sitzmann2019deepvoxels, lombardi2019neural}. Most of these works focus on static scenes \cite{mildenhall2020nerf, sitzmann2019deepvoxels}. \cite{lombardi2019neural} provides a way to control dynamic scenes implicitly via learning and traversing a latent space. Our work can be thought of as an extension of volume-based approaches on NVS from static scenes to poseable humans with explicit controls. Our representation builds on~\cite{lombardi2019neural}; they use volume warping fields to improve volume resolution when reconstructing scenes captured with a multi-view rig, while we leverage this idea to constrain the solution space for reconstructing poseable humans in single-view videos.

\textbf{Human mesh reconstruction} Human mesh reconstruction has drawn much attention recently and achieved significant success \cite{alldieck2019tex2shape, saito2019pifu, saito2020pifuhd, alldieck2019learning, huang2020arch, lazova2019360, varol2018bodynet, alldieck2018video, weng2019photo, zhu2020reconstructing}. Some of them focus on rebuilding mesh only \cite{saito2019pifu, saito2020pifuhd, varol2018bodynet, zhu2020reconstructing}, while others hook a rigged human model such as SMPL to the output \cite{alldieck2019learning, alldieck2019tex2shape, lazova2019360, huang2020arch, alldieck2018video, weng2019photo} to make it animatable. In particular, learning based methods need to work on synthetic data \cite{ saito2019pifu, saito2020pifuhd, varol2018bodynet, zhu2020reconstructing} or 3D scans \cite{alldieck2019tex2shape, alldieck2019learning, lazova2019360}, due to lack of ground truth in the wild data. For non-learning based method, \cite{alldieck2018video} solves it as an constrained optimization problem and \cite{weng2019photo} propose a novel warping method, to appropriately deform SMPL models to match input data. But both of them require the subject has either A-pose or fairly frontal pose so are hard to scale to general data in the wild. Different from previous works, our method directly learns from images in the wild in a self-supervised manner.

%% file: representation.tex
\section{Representation and Formulation} 
\label{sec:human_representation}

In this section, we describe our problem formulation and our representation for a posable volumetric model of a person.

% \section{Animatable Human Representation}

% \subsection{Problem Formulation}
\textbf{Problem Formulation:}  We take as input video frames $\{I_1, I_2, ..., I_N\}$ of a human subject performing various actions ``in the wild".  For each frame, using existing methods, we recover 3D body pose and camera parameters, resulting in a set of poses $\{\pose_1, \pose_2, ..., \pose_N\}$ and cameras $\{\cam_1, \cam_2, ..., \cam_N\}$. Our goal is to build an image synthesis function $\Iren(\pose, \cam)$ that renders the person in any pose $\pose$ from any viewpoint $\cam$. Thus, we aim to solve an optimization problem defined as: 

% The self-supervised approach of learning human rendering function from images can be formulated as an image synthesis problem as follows: given a set of images $\Gamma=\{I^1, I^2, ..., I^N\}$ capturing a human subject, with 3D body poses $P=\{ \rho^1, \rho^2, ..., \rho^N\}$ and cameras $\Phi=\{\phi^1, \phi^2, ..., \phi^N\}$, we want to find an image synthesis function $I^{syn}(\rho, \phi)$, taken $\rho$ and $\phi$ as inputs, to well reconstruct images in $\Gamma$. We then aim for solving an optimization problem:

\begin{equation}
\label{eq:min_img_synthesis}
    \argmin_{\Iren(\cdot)} \sum_{i=1}^N \Lagr(\Iren(\pose_i, \cam_i), I_i),
\end{equation}
where $\loss$ is the loss between $\Iren(\cdot)$ and $I$.

\textbf{Representation considerations:} Solution to Eq.~\ref{eq:min_img_synthesis}, requires the ability to modify both pose and camera view, while synthesizing the image. Most recent state-of-the-art methods that do not require camera rigs focus on one but not both of these.  For example, image-to-image translation methods~\cite{chan2019everybody, wang2018video} allow changing only 2D pose and do not scale well to 3D camera view changes, while methods like NeRF~\cite{mildenhall2020nerf} produce beautiful results for viewpoint changes around static scenes but do not handle dynamic scenes. 

Given recent success with implicit functions for scene and person modeling, we adopt a volumetric representation of the form $V(\mathbf{x}; \pose)$ that, for pose $\pose$, maps voxel $\x$ to $RGB\alpha$.  We can then synthesize an image as:
\begin{equation}
    \Iren(\pose,\cam) = R(V(\mathbf{x}; \pose),\cam)
\end{equation}
where $R$ produces a volume rendering of $V$ from viewpoint $\cam$.  Note that this representation corresponds to a diffuse scene approximation, i.e., no view-dependent reflection (an area of future work).

In principle, we could construct a deep network with weights $\theta$ to produce the volume directly, call it $V_\theta(\x; \pose)$.  While previous implicit networks have operated on 3D ($\x$) or 5D ($\x$ plus direction) and take as input many views of one static scene, $V_\theta(\x; \pose)$ operates on a $3+M$ dimensional domain, where $M$ is the number of pose parameters, typically $M>60$.  Without a very high capacity network and a prohibitive amount of training data (many views of many poses), such a network will struggle to produce high quality results, as we show later in the paper.

Instead, we decompose the problem into solving for a canonical $RGB\alpha$-volume and a motion weight volume; the motion weights are then used to warp the canonical volume based on pose.  This formulation builds on Lombardi~et~al.~\cite{lombardi2019neural}, who used warped template volumes to increase resolution and aid in motion representation when reconstructing subjects in a multi-view rig.  We adapt this approach to constrain the space of warps to follow a skeletal model to accommodate single-view videos in the wild. 

\textbf{Canonical Volume:} We define a canonical pose $\pose_c$ (e.g., the human ``T-pose'') and corresponding canonical $RGB\alpha$-volume $V^c$ for that pose. Given any other pose $\pose$, then, we warp $V^c(\x)$ to the target posed volume $V(\x; \rho)$:
\begin{equation}
\label{eq:warp_volume}
 V(\x; \pose) = V^c(T(\x; \pose))
\end{equation}
where $T$ encodes a volumetric warp function determined by pose.  Note that the canonical volume itself is just a function of $\x$, making it more tractable to recover.  With the warping function (described below), we can deform the canonical volume to match the input views; alternatively, we can conceptually think of the reverse direction as reposing the input views to the canonical space, to drive optimization of the $RGB\alpha$-volume function, akin to the more standard single-pose-multi-view case.

It is important to note that we have made a trade-off: we are assuming that the appearance of the subject is not pose-dependent.  As a result, strong shading and shadowing effects -- e.g., from directional lighting -- and wrinkles that may change as the person moves are not accurately reproduced and indeed make reconstruction more challenging.  Later, we design loss functions that enable plausible reconstruction of people despite this limitation. 

\textbf{Motion Weight Volume:} Although the canonical volume is a function of $\x$, the transformation function $T$ is still a function of many variables (position and pose) and in its general form is still prohibitive to reconstruct.

Instead, following Lombardi~et~al.~\cite{lombardi2019neural}, we model the deformation as a form of volumetric linear blend skinning~\cite{lewis2000pose}.  In particular, we define the transformation function as:
\begin{equation}
\label{eq:weight_sum_approx}
   T(\x; \pose) = \sum_{i}{\weightnorm_{i}(\x; \pose) B_i(\x; \pose)},
\end{equation}
where
\begin{equation}
\label{eq:motion_basis_and_weights}
    B_i(\x; \pose) = A_i(\pose)\x + \mathbf{t}_i(\pose), \;\;
    \weightnorm_i(\x; \pose) =  \frac{w_i(B_i(\x; \pose))}{\sum_j{w_j(B_i(\x; \pose))}}.
\end{equation}
$B_i$ is the $i$th motion basis (corresponding to the $i$th bone in a skeletal representation), defined by pose-determined matrix $A_i(\pose) \in \mathbb{R}^{3 \times 3}$ and translation vector $\mathbf{t}_i(\pose) \in \mathbb{R}^{3 \times 1}$, which can be explicitly computed via the corresponding body part transformation from target pose $\pose$ to canonical pose $\pose_c$, and $w_i$ is a weight volume in canonical coordinates.  Note that unlike traditional linear blend skinning that warps a mesh vertex from a canonical pose to a target pose, this formulation operates volumetrically and maps from a target pose back to the canonical pose.  In addition, while \cite{lombardi2019neural} solves for the motion bases, in addition to the weights, we prescribe the motion bases through sekeletal transformations and focus only on recovering the weights.

In practice, we have $K$ motion bases, corresponding to the number of bones, typically $K>20$.  We add one more weight channel, corresponding to a background class (empty space), as discussed in Sec.~\ref{sec:implementation}, and then pack the weight volumes into one $K+1$-channel volume, $W(\x)$.  

\textbf{Mask Volume:}  The weighted-sum-of-motion-bases assumption is only valid for the subject (i.e., foreground) voxels.  We therefore need a volume mask $M(\x;\pose)$ to identity them. We could reconstruct this mask directly, but we are then back to the high dimensional problem.  A key observation, to remove  pose dependency, is that $M$ has an implicit relationship with $W$. Specifically, $M$ can be approximated as follows:
\begin{equation}
\label{eq:mask_volume}
    M(\x; \pose) = \sum_{i=1}^K{w_i}(B_i(\x; \pose)),
\end{equation}
The intuition is by summing across all motion weights for different body parts we get a reasonable approximation of how likely a voxel belongs to the subject. Notice we warp the voxel $\x$ via $B$ before sampling weights because $W$ is defined in canonical space.

\begin{figure*}
  \begin{center}
  \includegraphics[width=\textwidth]{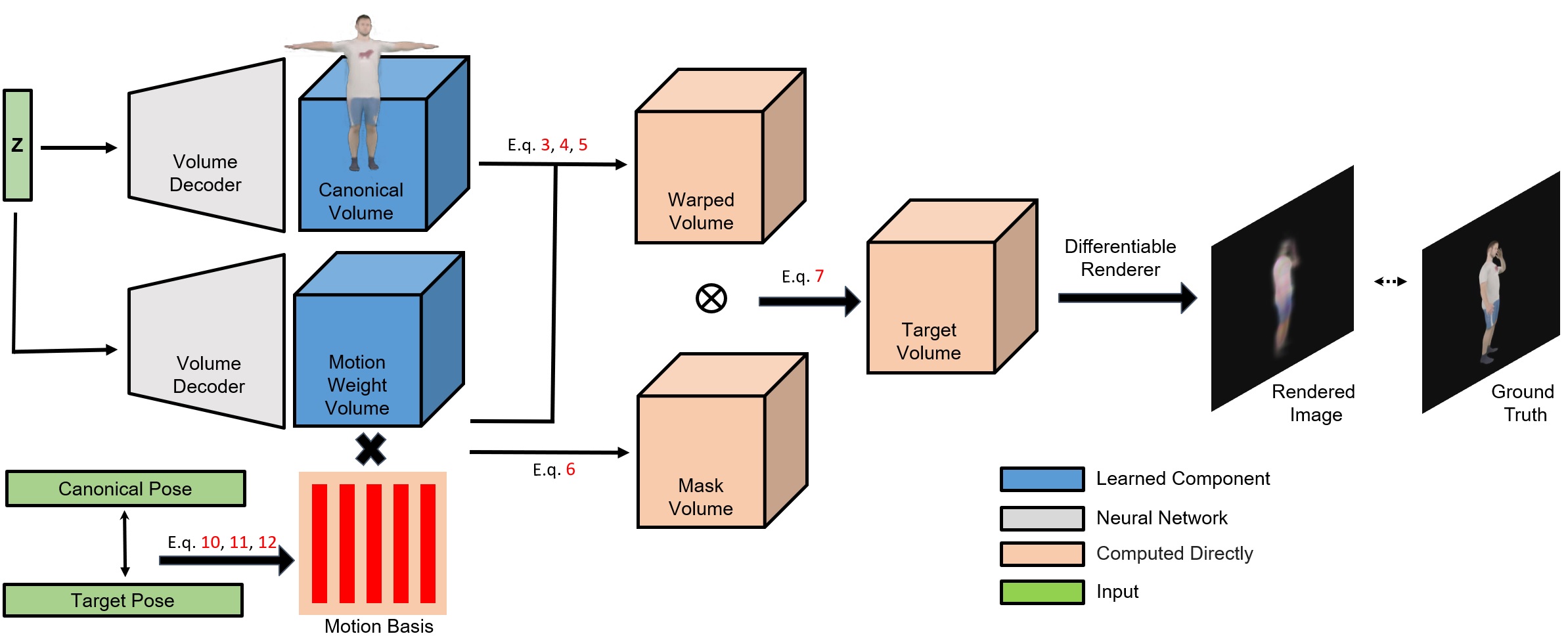}
  \end{center}
  \caption{Overview of our method. We begin with a constant vector $\mathbf{z}$ and produce a canonical volume and motion weight volume via neural networks. Then, combined with motion bases, derived from a predefined canonical and target pose, we explicitly deform the canonical volume to a warped volume and derive a mask volume, which is then used to mask out the warped volume to get a target volume. Finally, we render the target volume, and then optimize the loss between a rendered image and ground truth w.r.t the network parameters to get the best canonical and motion weight volumes.}
  \label{fig:overview}
\end{figure*}

Finally, the target posed volume $V(\x;\pose)$ can be computed by warping the canonical volume with $T$ and masking it out with $M$:
\begin{equation}
\label{eq:target_volume}
 V(\x; \pose) = M(\x; \pose) V^c(T(\x; \pose))
\end{equation}

{\bf Formulation with deep networks:} The goal now is to reconstruct the canonical volume $V^c(\x)$ and weight volume $W(\x)$ that best explain the input views.  Rather than directly solve for these volumes -- the classical approach -- we solve for parameters of deep networks that generate them from a single, fixed (and randomly assigned) latent variable $\z$: 
\begin{equation}
\label{eq:optimal_volume}
    V^c_{\theta_c} = g_{\theta_c}(\z), \;\;\;\;
    W_{\theta_w} = h_{\theta_w}(\z),
\end{equation}
We can now pose an optimization problem to solve for the network parameters:
\begin{equation}
\label{eq:min_final}
    \theta_c^*, \theta_w^* = \argmin_{\theta_c, \theta_w} 
    \sum_{i=1}^N \loss(R(V_{\theta_c,\theta_w}(\x,\pose_i),\cam_i), I_i),
\end{equation}
where $V_{\theta_c,\theta_w}(\x,\pose)$ is the masked, reposed canonical volume (Eq. \ref{eq:target_volume}) derived from the deep networks and, again, $R$ is the volume rendering function.  Note that these networks are only used as a framework to reconstruct the canonical and weight volumes; after the network weights are optimized, these volumes are fixed, enabling synthesis of novel views/poses, and the networks are no longer used.

%% file: method.tex
\section{Network Architecture and Training}
\label{sec:implementation}

In this section, we will describe how we represent and solve Eq.~\ref{eq:min_final} with neural networks, including the network architecture, network components, and how we optimize over synthetic data and Internet videos.

\subsection{Network pipeline, motion bases, and rendering} 

Figure~\ref{fig:overview} shows an overview of our method. Specifically, with a predefined canonical pose, we compute motion bases between the canonical and target poses. Then, we simultaneously produce a canonical volume $V^c$ and motion weight volume $W$ through two neural networks that take a single, constant, randomly initialized latent vector $\z$ as input. Combined with the motion bases, we can explicitly deform the canonical to a warped volume (Eq.~\ref{eq:warp_volume}-\ref{eq:weight_sum_approx}) and derive a mask volume (Eq.~\ref{eq:mask_volume}), which will be used to mask out the warped volume to get a target volume (Eq.~\ref{eq:target_volume}). Finally, we render the target volume with a differentiable renderer to get a synthesized image, and minimize the loss between the synthesized and ground truth images w.r.t the network parameters to get the final volumes $V^{c}_{\theta^*_c}$ and $W_{\theta^*_w}$.

\textbf{Network architecture} We represent a canonical volume $V^c$ as a 3D voxel grid with grid size $128^3$ and 4 channels ($RGB\alpha$), and a motion weight volume $W$ with grid size $64^3$ and $K+1$ channels, where $K$ is the number of body parts. The constant input $\mathbf{z}$ is a 256-dimensional vector, initialized with a Gaussian distribution. The network $g_{\theta_c}$ producing $V^c$ begins with a MLP with LeakyRelu non-linearity to transform z and reshape it to a $1 \times 1 \times 1 \times 1024$ grid, and then concatenates 7 transposed convolutions with LeakyReLU non-linearity. The network $h_{\theta_w}$ producing $W$ has similar architecture but with 6 transposed convolutions. Additionally we apply \textit{softplus} to the output of $g_{\theta_c}$ and \textit{softmax} to $h_{\theta_w}$\footnote{See \ref{sec:implementation_details} to know why we apply \textit{softmax} to $h_{\theta_w}$.}  to guarantee non-negative values.

\textbf{Motion bases:} We use the SMPL skeleton framework to derive motion bases. In SMPL \cite{loper2015smpl}, a body pose is defined as a tuple, $(\joints, \jangles)$, where $\joints \in \mathbb{R}^{K \times 3}$ includes $K$ standard 3D joint locations, and $\jangles \in \mathbb{R}^{3K \times 1} = [\jangle_0^T, ..., \jangle_K^T]^T$ defines K axis-angle representations, $\jangle \in \mathbb{R}^{3 \times 1}$, of the relative rotation of body part w.r.t to its parent in the kinematic tree. Following the same convention, we represent our predefined canonical pose $\pose_c$ as $(\joints_c, \jangles_c)$, and denote a target pose $\pose_t$ as $(\joints_t, \jangles_t)$. The transformation $F_k$ of body part $k$ from $\pose_t$ to $\pose_c$ is then:

\begin{equation}
\label{eq:motion_basis_f}
    F_k(\joints_c, \jangles_c, \joints_t, \jangles_t) = G_k(\joints_c, \jangles_c)G_k(\joints_t, \jangles_t)^{-1} ,
\end{equation}

\begin{equation}
\label{eq:motion_basis_g}
    G_k(\joints, \jangles) = \prod_{i \in P(k)}
        \begin{bmatrix}
            R(\jangle_i) && \joint_i \\
            0 && 1
        \end{bmatrix},
\end{equation}

where $R(\jangle_i) \in \mathbb{R}^{3 \times 3}$ is a rotation matrix by applying \textit{Rodrigues formula} to $\jangle_i$, $\joint_i \in \mathbb{R}^{3 \times 1}$ is the $i$th joint center location in $\jangles$ defined in its local space, and $P(k)$ is the ordered set of parents of joint $k$ in kinematic tree.

The motion basis of body part $k$, defined as a tuple $(A_k, \mathbf{t}_k)$, is then:

\begin{equation}
\label{eq:motion_basis}
    \begin{bmatrix}
            A_k && \mathbf{t}_k \\
            0 && 1
    \end{bmatrix} = 
    F_k(\joints_c, \jangles_c, \joints_t, \jangles_t)
\end{equation}

\textbf{Differentiable Volume Renderer:} Following \cite{lombardi2019neural}, we formulate the volume rendering as a front-to-back additive blending process, accumulating color and $\alpha$ while ray-marching through the volume, stopping whenever accumulated $\alpha=1$. 

The volume renderer produces a color image $I$ and $\alpha$ image which can be composited per pixel over a background image, $I_{\text{bg}}$, to get the final output:

\begin{equation}
    \Iren = I + (1 - \alpha) I_{\text{bg}}
\end{equation}

Our loss functions will ultimately compare $\Iren$ to ground truth images.  The background for synthetic ground truth images are known trivially, whereas, for real imagery, we mask out the background after segmentation and replace with a known color before comparing to $\Iren$ composited over that same color.

\subsection{Learning from data}

We now describe how we learn the network weights to generate canonical volumes $V^c$ and motion weight volumes $W$ using synthetic and real data.

\subsubsection{Synthetic data}

For synthetic data, we render a SMPL-based character model in a variety of poses with virtual cameras placed evenly on a sphere using an emissive texture. To simulate the single-view scenario (not multi-view per pose), each pose will be rendered only once by one view, randomly selected from training cameras; we hold out a set of images for testing. More details can be found in Sec. \ref{sec:experiment}.

\textbf{Canonical pose:} Without loss of generality, we define the canonical pose as $\pose_c$ as $(\joints_c, \jangles_c)$, with $\joints$ set to the default SMPL joint positions, and $\jangles_c$ is a all-zeros vector corresponding to a T-pose. Note that we intentionally omit the T-pose from the training data.

\textbf{Loss function:} For this synthetic case, we use a simple L2 loss summed across pixels:
\begin{equation}
    \Lagr = \| \Iren(\pose_i,\cam_i) - I_i \|^2_2,
\end{equation}
where $I_i$ is the ground truth image and $\Iren$ is the rendered image for corresponding pose $\pose_i$ and camera $\cam_i$.

\begin{figure*}
  \centering
  \includegraphics[width=\textwidth]{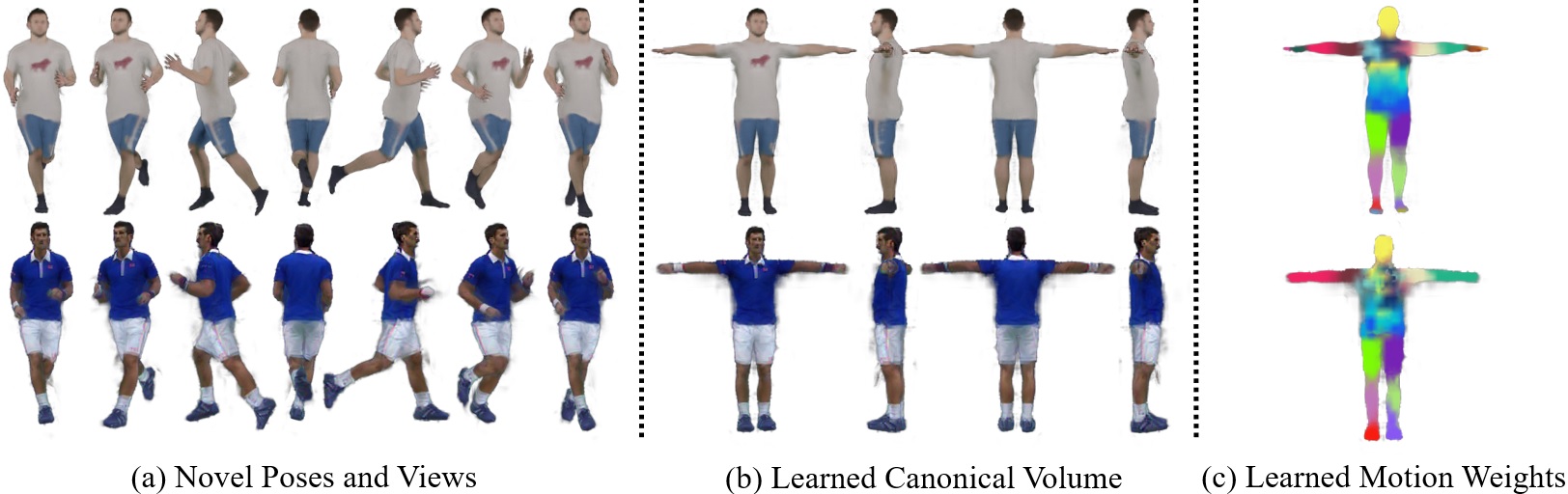}
  \caption{Results of Vid2Actor for synthetic (top row) and real data (bottom row, result on video of Novak Djokovic).}
  \label{fig:results}
\end{figure*}

We found that this simple loss worked reasonably well for the synthetic case (see the 1st row of Fig. \ref{fig:results}).  Rendering unseen poses to novel views produces results very similar to ground truth (see Fig. \ref{fig:vs_baseline} and Table. \ref{table:vs_baseline}). We find that the canonical volume is similar to what we get when training on multi-view data of just that pose, and the distribution of motion weight volume aligns well with the body parts. We believe the success comes from the fact that the pixels across different images can be well aligned via learned warping and thus succeed in constraining the synthesis of the canonical volume.

\subsubsection{Videos ``in the wild"}
\label{sec:wild_data}

Videos ``in the wild'' introduce more challenges, as 3D body pose and camera estimation, as well as person segmentation, even with state-of-the-art methods, yield imperfect results, particularly when working with Internet videos where extreme conditions are common. We handle this problem in two complementary ways: (1) automatically filter out frames with likely pose/segmentation errors and (2) adjust the loss function to recover detail despite misregistrations.

\textbf{Data preprocessing:} We run SPIN \cite{kolotouros2019learning} to get camera and 3D body pose per frame. Then,  Pointrend \cite{kirillov2020pointrend} is applied for person segmentation. Additionally, we estimate 2D body pose using OpenPose \cite{cao2018openpose}, to validate the correctness of 3d pose estimation.  We crop and scale each image so that: (1) its resolution is 512x512, (2) the subject is centered, and (3) the longer side of subject bounding box is 400 pixels; as a result, the subjects have approximately the same size across all images, which aids in training.

\textbf{Data filtering:} We filter out those images with potential estimation errors. Specifically, we drop an image if it has (1) partial body in frame, as 3D pose estimation usually fails in the case, (2) inconsistency between 2D and 3D pose, which implies incorrect pose estimation, or (3) inconsistency between pose and segmentation.

Specifically, we check (1) by testing if OpenPose detects ankles and wrists and (2) by projecting 3D pose to the image plane and comparing it with 2D pose result from OpenPose.

For (3), we compare the detected person mask, $I_{\text{pm}}$, to the silhouette image, $I_{\text{sil}}$, rendered with the posed SMPL model given the estimated 3D pose. Specifically, we filter out an image if:
\begin{equation}
        \frac{|I_{\text{pm}} \cap I_{\text{sil}}|}{|I_{sil}|} < 0.95,
\end{equation}
The intuition is that $I_{sil}$ should be completely covered by $I_{pm}$ if $I_{pm}$ is correct, since the SMPL model is semi-nude.

\textbf{Canonical pose:} The SPIN pose estimator recovers joint locations $J_{\beta}$ per frame, where $\beta$ is the estimated shape parameter vector for the frame.  We compute $\bar{\beta}$ as the average of the shape vectors across frames that were not filtered out and set the canonical joint locations to $\joints_c = \joints_{\bar{\beta}}$.  As with the synthetic case, we set the joint angles $\jangles_c$ to the zero vector, corresponding to the T-pose.

\textbf{Loss function:} 
After the filtering process, the remaining frames generally have reasonable estimated poses and segmentations, but they are still not precise. In addition, even with perfect information, pose- and view-dependent effects such as wrinkles, shadows, or specular highlights in real data violate our pixel consistency assumption.  The resulting mis-registrations and pose+view-dependent effects will result in averaging out of details, and worse, when using a simple L2 loss.

To mitigate these problems, we resort to perceptual loss, specifically VGG loss, because it (1) compares high-level features rather than low-level pixels, making it robust to appearance changes; (2) is convolutional, providing some robustness to small translation shifts, which helps balance estimation errors; (3) it can create high frequency details in image synthesis~\cite{chen2017photographic}. In practice, we  add L1-norm loss to encourage learning convergence, and only apply VGG loss on foreground regions (identified using estimated masks).

The results are significantly improved after replacing L2 with L1 + VGG loss (see Fig. \ref{fig:ablation_study}). However, the face region, critically important, can still be distorted. Thus, we add a GAN loss specific to head regions, learned from the input data, to aid in recovering facial details.

Our final loss function is then: 
\begin{equation}
     \Lagr = \Lagr_{VGG} + \lambda_1\Lagr_{L1} + \lambda_2\Lagr_{FaceGAN},
 \end{equation}
with $\lambda_1$ and $\lambda_2$ set to 0.1 and 0.001  experimentally.

\begin{figure*}
  \centering
  \includegraphics[width=\textwidth]{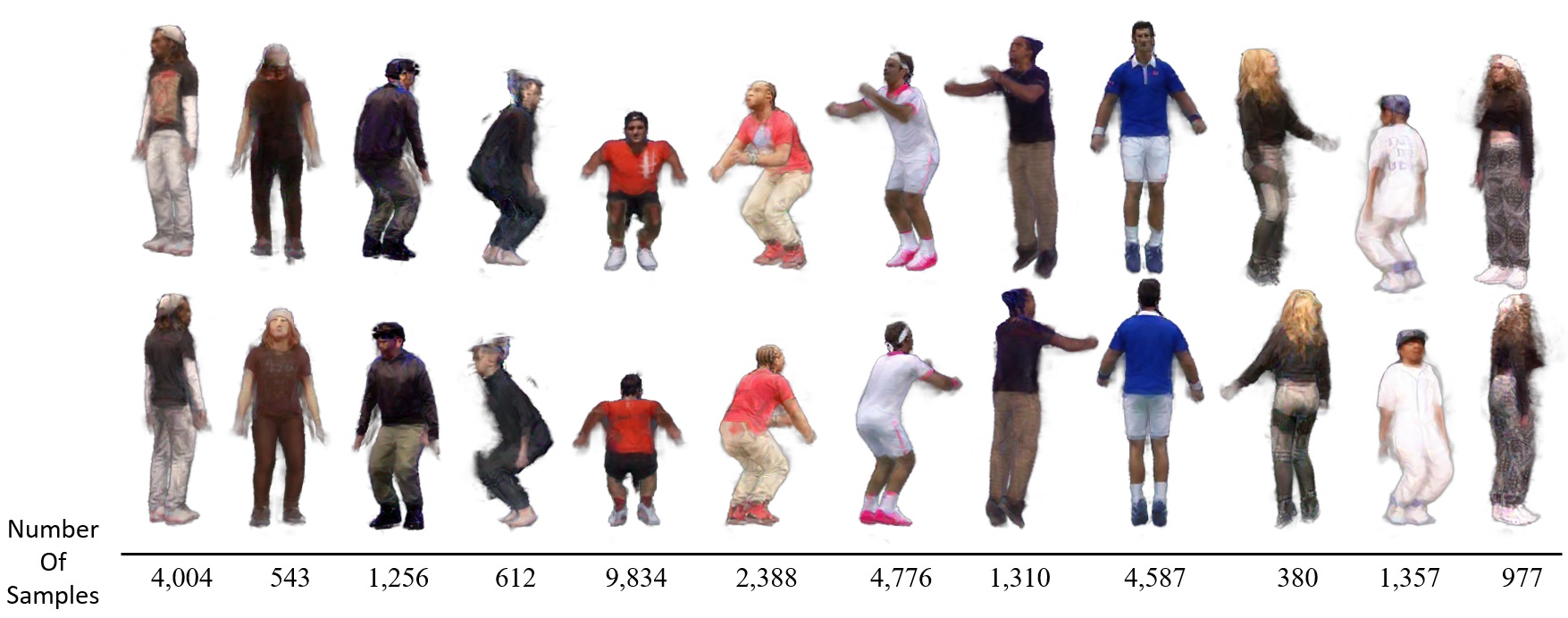}
  \vspace{-4ex}
  \caption{Motion Re-targeting. We apply jumping to 12 learned human models. Below the model we show the number of frames used for training. In general, more frames lead to sharper results.}
  \label{fig:retargetting}
  \vspace{-2ex}
\end{figure*}

\subsection{Implementation Details}
\label{sec:implementation_details}

\textbf{FaceGAN loss:} We train a 3-scale-patchGAN discriminator with an LSGAN loss \cite{wang2018high}. We crop out a 64x64 region centered around the head in each image to provide training data. 

\textbf{Random background:} For the data-in-the-wild scenario, if we use a single background color during training, we find it will be incorrectly used as part of the foreground, yielding noisy results. To avoid this problem, we randomly assign a random (solid) background color to each training image and corresponding rendering image.

\textbf{Additional background channel in $\mathbf{W}$:} A voxel in $W$ is either associated primarily with one or two body parts or is in the background (empty space). We add one more channel to $W$ for this background category. We then task the network with producing a volume with channel size $K+1$ (for $K$ bones) and apply softmax to the result:
\begin{equation}
    W(\mathbf{x}) = \text{softmax}(\Wtilde(\mathbf{x})),
\end{equation}
where we denote the network output as $\Wtilde$.  The intuition is via \textit{softmax} and the background channel, we can explicitly enforce sum-to-one constraint to voxels in $W$. 

\textbf{Learning $\mathbf{\Delta W}$:} To make it easier to decouple canonical and motion weight volumes, we task the motion weight network $h_{\theta_w}$ with generating $\Delta W$ instead, the difference between $\Wtilde(\x)$ and the logarithm of a pre-computed weight volume $W_G$.  $W_G$ consists of an ellipsoidal Gaussian around each bone (one Gaussian per bone weight channel) specifying rough body part locations in the canonical pose $\pose_c$.  Thus, before applying softmax, we have:
\begin{equation}
    \Wtilde(\x) = \Delta W_{\theta_w}(\x) + \text{log} W_G(\x)
\end{equation}
where $\Delta W_{\theta_w} = h_{\theta_w}$, and the background weight in $W_G$ is set to one minus the sum of all the bone weights in that volume.  Note that we apply the log to the Gaussian weight volume, which will then be exponentiated again when taking the softmax.

\textbf{Training details:} We use Adam \cite{kingma2014adam} to optimize the loss function with fixed learning rate $10^{-4}$ and batch size 2. The training takes roughly take 50,000 to 200,000 iterations to reach convergence, depending on the data set.

%% file: experiment.tex
\section{Experiments}
\label{sec:experiment}
\vspace{-1ex}

In this section we compare our results to related baseline methods, show ablation studies, and results. {\bf Please refer to the supplementary video for dynamic results\footnote{\url{https://youtu.be/Zec8Us0v23o}}. }

\textbf{Comparison with baselines:} We compare with two baseline methods to justify our design choices in Sec.~\ref{sec:human_representation}. We use synthetic data for the comparison, as it provides clean training signals to all the methods.

\textit{pix2pix3D}: We modify pix2pixHD \cite{wang2018high} to allow explicit pose and view controls. We use a skeleton map, $\kappa$, as an input, where we project 3D body pose $\pose_i$ to the image plane of camera $\cam_i$.  We represent $\kappa$ as a stack of layers, where a layer is an image representing a body bone. To keep 3D information in $\kappa$, we rasterize each bone into its layer by linear-interpolating the $z$-values of its endpoints.

\textit{Direct method (PoseVolume)}: We condition a volume generation network $V_{\theta}(\x;\pose)$ directly on 3D body pose $\pose$ as explained in Sec. \ref{sec:human_representation}. This means the network has to be able to estimate target volumes directly. The network architecture is same as we used in producing a canonical volume, except the MLP takes 72-dimensional pose vector as input.

\textbf{Synthetic data:} To generate synthetic data, we get a character model from the \textit{people-snapshot} data set~\cite{alldieck2018video} and use CMU mocap data~\cite{cmumocap2007} to repose it. We render the posed subject with 144 cameras spaced evenly on a sphere, where we randomly hold out 10\% of poses and 10\% cameras for testing. In addition, we use a smaller sphere radius for cameras used in testing (radius 1.5 for testing and 2.25 for training), to provide more significant view changes. To simulate the single-view scenario (vs. many views of one pose), each body pose will be only rendered once with one view, randomly selected from the training cameras. In total, we have 7653 training and 851 testing images. The two baselines and our method use the same training and testing data.  We use Adam to train all models until they are converged.

\begin{figure}[H]
  \centering
  \vspace{-2ex}
  \includegraphics[width=\linewidth]{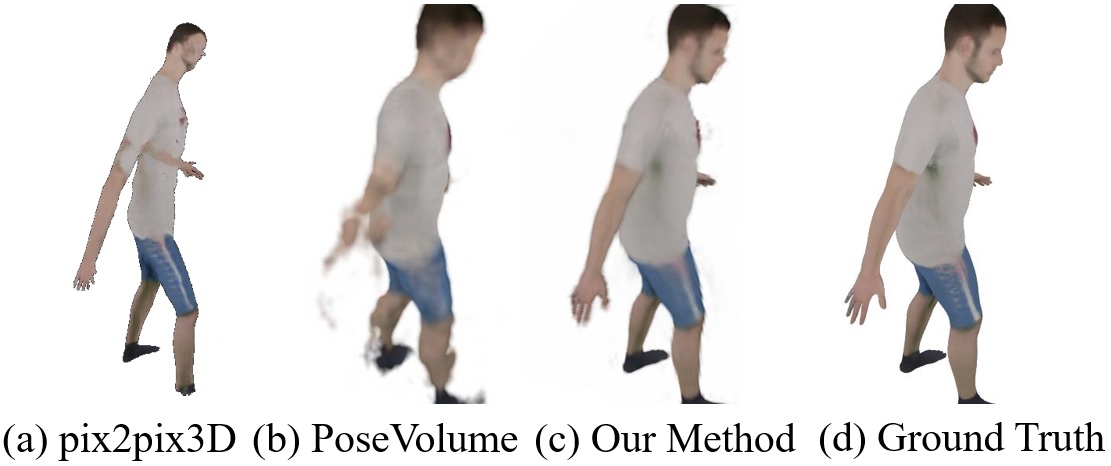}
  \vspace{-4ex}
  \caption{Qualitative comparison with baselines.}
  \label{fig:vs_baseline}
  \vspace{-2ex}
\end{figure}

\textbf{Comparison results:} The results of our method in comparison to the other two baselines are shown in Fig.~\ref{fig:vs_baseline}. Our method produces higher quality renderings than the baselines. The quantitative comparison is in Table. \ref{table:vs_baseline}. 

\begin{table}[H]
\vspace{-1ex}
\begin{center}
\begin{tabular}{ |c|c|c| }
\hline
 & PSNR $\uparrow$ & Mean Square Error $\downarrow$ \\
\hline
\hline
pix2pix3D \cite{wang2018high} & 17.54 & 1143.65 \\
PoseVolume &  23.98 &  351.99 \\ 
Our Method &  \textbf{30.13} & \textbf{69.22} \\ 
\hline
\end{tabular}
\end{center}
\vspace{-3ex}
\caption{Quantitative results compared to two baselines. } \label{table:vs_baseline}
\vspace{-2ex}
\end{table}

\textbf{Internet videos:} Our test videos were all collected from the Internet. We ran our method on videos with subjects performing unconstrained activities such as playing tennis or dancing. The video length varied between a few minutes to 2 hours. After pre-processing and filtering, the number of training frames ranged from about 300-10,000 frames. 

\begin{figure}[H]
  \centering
  \includegraphics[width=\linewidth]{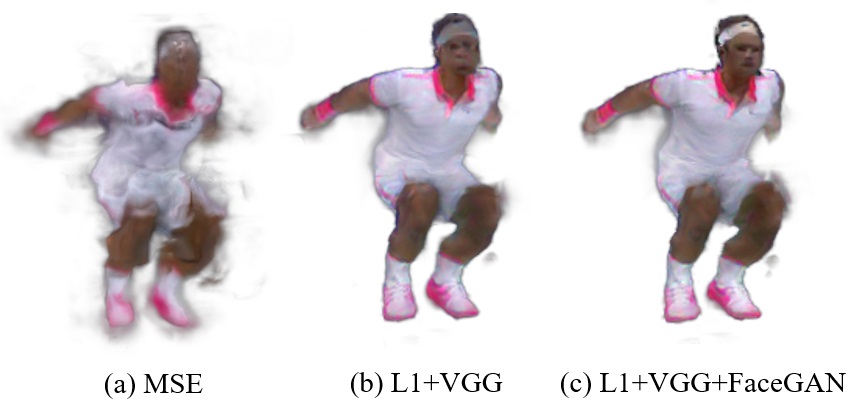}
  \vspace{-3ex}
  \caption{Qualitative ablation study where we evaluate how different loss functions affect the  rendering.  Notice the face and body details in (c), in comparison to (a) and (b).}
  \label{fig:ablation_study}
  \vspace{-2ex}
\end{figure}

\textbf{Qualitative ablation study:} We compare different loss functions and their performance on videos in the wild. 
Fig.~\ref{fig:ablation_study} shows how the rendering quality improves as the perceptual losses are introduced.

\begin{figure}[H]
  \centering
  \includegraphics[width=\linewidth]{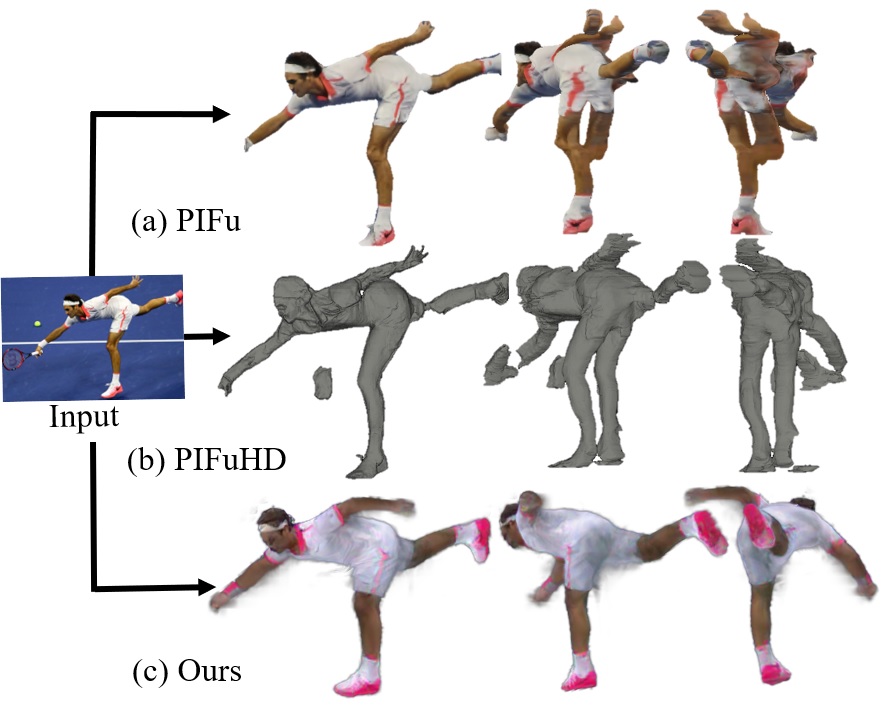}
  \caption{Bullet-time rendering and comparisons: (a) PIFu \cite{saito2019pifu}, (b) PIFuHD \cite{saito2020pifuhd}, and (c) ours. \scalebox{.75}{(Getty images/Maddie Meyer)}}  
  \label{fig:vs_pifu}
\end{figure}

\textbf{Results:} We show our results for Internet videos in several figures (see Fig. \ref{fig:teaser}, \ref{fig:results}, \ref{fig:retargetting}, \ref{fig:ablation_study}, \ref{fig:vs_pifu}).  The learned model is able to generalize to unseen poses and views, and the learned canonical and motion weight volumes (see Fig. \ref{fig:results}) perform similarly well for both Internet and synthetic data. In general, more frames lead to sharper results  (see Fig. \ref{fig:retargetting}). But even trained with a relatively small number of frames ($<$1,000), the model still performs reasonably well, aided by the decomposition into canonical volume and motion weights.  We note that the reconstructions tend to have blobby hands; we do not have reliable 3D hand pose estimations and thus don't register them well across views.

\textbf{Motion Re-targeting:} To enable motion re-targeting, we simply apply mocap data to learned human models by feeding 3D poses. The re-targeting results can be found in Fig. \ref{fig:teaser}, \ref{fig:results}, and \ref{fig:retargetting}. In particular, Fig. \ref{fig:retargetting} shows the result where we enforce jumping sequence to all of our models.

\textbf{Bullet-time rendering:} Since our model supports explicit pose and view controls, achieving bullet-time rendering from single images is straightforward. We first run 3D human pose estimation, and then render the posed model, with the detected body pose, by rotating the detected camera around the subject. 

We compare our method with PIFu \cite{saito2019pifu} and PIFuHD \cite{saito2020pifuhd} in bullet-time rendering as they provide alternatives to achieve the same effect but via 3D reconstruction. To get the best performance, we provide them the perfect person masks labeled manually. Note that although PIFu supports multiple-views, it requires the subject to hold the same pose across different views (or a multi-view rig), which does not hold in our case. We thus apply PIFu's single view mode.

The results are shown in Fig. \ref{fig:vs_pifu}. We found both PIFu and PIFuHD generally work well on fairly frontal poses, but not as well on others. In our case poses may be quite diverse. In contrast, our method performs well with significant pose changes, and preserves texture details in various views. 

%% file: conclusion.tex
\section{Discussion}
 \textbf{Limitation:} Our model quality depends on the diversity of poses and views in the training data. A wider variety of data leads to sharper results and better reconstruction of unseen configurations. As noted in the introduction, we assume the appearance of the subject is neither view-dependent nor pose-dependent, which may not be hold in reality. We mitigate these limitations to some extent by introducing perceptual losses, leading to plausible, if not physically correct, reconstructions. Modeling pose-dependent lighting effects and more complex shape deformations are exciting areas for future work. As a promising example, with a light stage\cite{guo2019relightables}, \cite{meka2019deep, Meka:2020} learn to synthesize one-light-at-a-time (OLAT) images that enables relightable videos. In addition, progress with MLP networks for implicit scene reconstruction~\cite{mildenhall2020nerf, sitzmann2020implicit} suggests a path toward increased resolution beyond voxel grids.
 
 %For example, \cite{meka2019deep, guo2019relightables, Meka:2020} leverage a light stage to approximate a mapping function from gradient color images to one-light-at-a-time (OLAT) images to enable relightable videos. In addition, our resolution is in part limited by using a voxel grid; implicit MLP networks~\cite{mildenhall2020nerf}, while currently slower to render, hold great promise for increasing effective resolution, another area for future work.
 
\textbf{Summary:} We propose a novel but simple animatable human representation, learned from only image observations, that enables person synthesis from any view with any pose. We validate it on both synthetic data and Internet videos, and demonstrate applications such as motion re-targeting and bullet-time rendering.  Our approach offers a new way to enable free-viewpoint animatable person reconstruction and rendering, providing an alternative to mesh-based representations.

\textbf{Ethics Statement:} This algorithm is intended as a step toward CGI effects for entertainment, e.g., sports visualization, and facilitating VR/AR applications such as telepresence.  That said, technology that enables re-animating and re-rendering humans could be used for creating false depictions.  Public or commercial deployment should be done with care, e.g., by watermarking the imagery.  At the same time, our methods will be in the open and made available for counter-measure analysis.

{\bf Acknowledgements} This work was supported by the UW Reality Lab, Facebook, Google, Futurewei, and Amazon. We thank all of the photo owners for allowing us to use their photos; photo credits are given in each figure.
% We believe the representation not only provides a convincing alternative, different from mesh based representation, to capture and render humans, but also, with its wild setup, suggests a pathway to build up practical VR/AR applications such as telepresence that benefits the society and brings people together around the world, amid the historic pandemic.

%\textbf{Acknowledgements:} This work was supported by the UW Reality Lab, Facebook, Google, Futurewei, and Amazon. We thank all of the photo owners for allowing us to use their photos.